\documentclass{article}

\usepackage{microtype}
\usepackage{graphicx}
\usepackage{subfigure}
\usepackage{booktabs} 

\usepackage{hyperref}
\usepackage{epsfig}
\usepackage{mahnigCEC}
\usepackage{url}
\usepackage{enumerate}
\usepackage{amsmath}
\usepackage{amsfonts}
\usepackage{tikz}

\usepackage[normalem]{ulem}
\usepackage{comment}

\usepackage{hyperref}

 \usepackage[accepted]{icml2024}

\usepackage{amsmath}
\usepackage{amssymb}
\usepackage{mathtools}
\usepackage{amsthm}

\usepackage[capitalize,noabbrev]{cleveref}

\theoremstyle{plain}
\newtheorem{theorem}{Theorem}[section]
\newtheorem{proposition}[theorem]{Proposition}

\theoremstyle{definition}

\theoremstyle{remark}
\newtheorem{remark}[theorem]{Remark}

\usepackage[textsize=tiny]{todonotes}

\newcommand{\R}{\mathbb{R}}

\newcommand{\norm}[1]{\left\Vert#1\right\Vert}
\newcommand{\inner}[2]{\left\langle{#1},{#2}\right\rangle}

\newcommand{\te}{\textstyle}

\icmltitlerunning{Submission and Formatting Instructions for ICML 2024}

\begin{document}

\twocolumn[

\icmltitle{Optimal synthesis embeddings}



\icmlsetsymbol{equal}{*}

\begin{icmlauthorlist}
\icmlauthor{Roberto Santana}{equal,ehu}
\icmlauthor{Mauricio Romero Sicre}{equal,ufba}
\end{icmlauthorlist}

\icmlaffiliation{ehu}{Department of Computer Science and Artificial Intelligence, University of the Basque Country (UPV/EHU), San Sebastian, Spain.}
\icmlaffiliation{ufba}{Instituto de Matemática e Estatística, Universidade Federal da Bahia, Salvador, Bahia, Brazil}

\icmlcorrespondingauthor{Roberto Santana}{roberto.santana@ehu.eus}
\icmlcorrespondingauthor{Mauricio Romero Sicre}{msicre@ufba.br}

\icmlkeywords{sentence embeddings, text data augmentation, sentence classification, NLP}

\vskip 0.3in
]



\printAffiliationsAndNotice{\icmlEqualContribution} 

\begin{abstract}
 In this paper we introduce a word embedding composition method based on the intuitive idea that a fair embedding representation for a given set of words should satisfy that the new vector will be at the same distance of the vector representation of each of its constituents, and this distance should be minimized. The embedding composition method can work with static and contextualized word representations, it can be applied to create representations of sentences and learn also representations of sets of words that are not necessarily organized as a sequence. We theoretically characterize the conditions for the existence of this type of representation and derive the solution. We evaluate the method in data augmentation and sentence classification tasks, investigating several design choices of embeddings and composition methods. We show that our approach excels in solving probing tasks designed to capture simple linguistic features of sentences.   
\end{abstract}

\section{Introduction}

In machine learning, words embeddings are extensively applied as an efficient representation for addressing a variety of tasks in natural language processing (NLP)  \cite{Iyyer_et_al:2014,Mikolov_et_al:2013b,Turian_et_al:2010}, task such as  analogy resolution \cite{Mikolov_et_al:2013,Mikolov_et_al:2013a}, sentence similarity estimation \cite{Conneau_et_al:2017},  and word-vector based clustering \cite{Zhai_et_al:2016}. Word embeddings can also be applied within the so-called  \emph{corpus centered approach} for investigating  gender and ethnic bias in the language \cite{Bolukbasi_et_al:2016,Garg_et_al:2018,Gonen_and_Goldberg:2019}, unearthing analogies in the literary discourse \cite{Heuser:2017}  and quantifying semantic change over time  \cite{Hamilton_et_al:2016}.


 
The question of how to design effective compositions of word embeddings able to capture the semantics of a phrase has been extensively investigated since the publication of the initial works showing that analogies could be solved by using word vector arithmetic's \cite{Mikolov_et_al:2013b}, and the following theoretical analyses of these findings \cite{Allen_and_Hospedales:2019,Arora_et_al:2016}. Particular emphasis has been put on the design and analysis of sentence representations that capture rich semantic information  \cite{Cer_et_al:2018,Li_et_al:2022}. 


 
 The problem we address in this paper is, given a set of words and their corresponding word-vectors, create (or define) its representation (word-vector) as a vector that is as close as possible, and at the same distance, of all the given word vectors. Crucially, the problem we deal with is more general than the sentence embedding representation problem, since a set of words does not imply a particular order of these words.  
 
 Having a representation equidistant from all given word vectors can be practical in situations where the goal is creating a single representation that synthesizes semantic attributes of different words. In this paper we present two scenarios to evaluate this type of representation: 1) Word data augmentation for word classification, where we create a word-vector that is at the same distance of embeddings corresponding to $k$ words in the same class; and 2) Sentence embedding creation for sentence classification, where we use all word embeddings of the words  in the sentence to create the equidistant vector. 
 
 


 In addition to introduce a  new method for word composition  useful for solving machine learning tasks, the method we propose can be seen through the prism of \emph{fair} embedding composition. Based on the generally accepted assumption that distance between word vectors can be associated to semantic similarity,  the solution we propose for embedding composition can be interpreted as being \emph{fair} for all its constituents, in terms of having a meaning equally similar to each of them. This can be of application when constructing a neutral semantic representation for two or more gender or ethnic variations of the same word. Furthermore, by minimizing the distance of this representation to  all the given vectors, we also aim for a vector that encodes a meaning as close as possible to the ones of the original words. 

   We start by discussing conditions for existence of vectors equidistant from a set of word vectors and deriving methods for computing these representations. We also derive the solution for a related word embedding synthesis problem. In the experimental part of the paper we propose the use of the introduced method for data augmentation and a diverse set of sentence classification tasks. We compute the results in terms of classification accuracy, comparing the introduced method with previous algorithms used for word embedding composition. 
   


   We summarize  our contributions as following:
   \begin{itemize}
    \item We theoretically characterize the conditions for the existence of an $n$-dimensional word vector which is at the same distance from $N$ other word vectors, and derive solutions to find this set of vectors, and among them the vector at minimum distance.
    \item We prove the conditions for the existence of a solution, and derive this solution, for the problem of  finding a word vector which is at given distances from $N$ other vectors.  
    \item We conduct extensive experiments to evaluate the performance of the word vector composition method in data augmentation and sentence classification tasks, using different types of word embeddings and comparing to other approaches for word embedding composition. 
   \end{itemize}

 \section{Related work}

 \subsection{Composed word-vectors and sentence embeddings}  



   The idea of producing new semantic information from the combination or composition of word vectors have arisen in different NLP applications. For example, different types of simple algebraic operations on word embeddings have been proposed to solve the word analogy task \cite{Blacoe_and_Lapata:2012,Levy_and_Goldberg:2014,Mikolov_et_al:2013,Vylomova_et_:2015}. Another type of composition is the one used for constructing sentence embedding representations.  In these cases, the simplest and most common strategy is to compute the average of all word embeddings in the sentence. This approach can be applied to context-independent \cite{Mikolov_et_al:2013a,Pennington_et_al:2014,Joulin_et_al:2016} and context-dependent word representations \cite{Devlin_et_al:2018,Liu_et_al:2019a}. 
   
   Several authors  have proposed methods that aim to compute a unique sentence representation that is not a simple combination of word embeddings \cite{Cer_et_al:2018, Conneau_et_al:2017, Conneau_et_al:2018}.  In \cite{Cer_et_al:2018}, two strategies for creating sentence embeddings are proposed, the transformer-based approach computes a context-aware representation of words by solving multiple NLP tasks, and then the sentence representation is calculated as the element-wise sum of the representations at each word position. In the deep neural network approach \cite{Cer_et_al:2018}, the input embeddings for words and bi-grams are first averaged together and then passed through a feedforward deep neural network.
   
   While the focus of the approaches presented in \cite{Cer_et_al:2018} and in other papers has been on the identification of the neural network models and the ML learning tasks needed to learn effective sentence representations, our goal in this paper is the conception of tools instrumental to create an effective representation from the given word embeddings, and possibly additional prior information such as distances, weights, etc., relevant for computing the global representation. 

 \subsection{Evaluating sentence representations} 

   There have been considerable effort in the creation of benchmarks to evaluate the properties encoded by sentence representations. The yearly SemEval tasks  \cite{Agirre_et_al:2012,Agirre_et_al:2013,Agirre_et_al:2014,Agirre_et_al:2015,Agirre_et_al:2016} have contributed to the creation of testing benchmarks and in this work we employ this benchmark to evaluate the feasibility of our method to create sentence embeddings. In addition, we employ a number of probing tasks.  A probing task \cite{Adi_et_al:2016,Conneau_et_al:2017,Shi_et_al:2016} is a classification problem that serves to evaluate what linguistic properties are encoded in sentence embeddings.



 \subsection{Enriching, using and evaluating semantic spaces}

  Another research direction related to our research includes proposals that design methods to enrich the embedding space with additional elements or other type of information.   Several works have considered how to enrich the embedding spaces with semantic information.  These include both, supervised and unsupervised methods. One of the  unsupervised approaches proposes transformations of the embedding spaces to adapt them to particular tasks.  In retrofitting \cite{Faruqui_et_al:2014}, information external to the embedding space (e.g., semantic lexicons) is used to encourage semantically related words to have similar vector representations. 

  Another type of supervised transformations are ultradense embeddings \cite{Rothe_et_al:2016,Dufter_and_Schuetze:2019}, in which the original embedding space is reordered concentrating the information relevant to a task in an ultradense subspace. This space provides a higher quality representation for the task.

\section{Notation} 


{We will denote by $<\cdot,\cdot >$ the standard inner product in $\R^n$ and by $\|\cdot\|$ the corresponding Euclidean norm.
Given a subspace $S$ of $\mathbb{R}^n$ and a vector $x\in\mathbb{R}^n$, we denote the projection of $x$ in $S$ by $P_S(x)=\underset{v\in S}{\mbox{argmin}}\norm{x-v}$. 
We will say that a set of linear independent vectors is l.i. and a set of linear dependent vectors is l.d.. We will denote by $\bf e\in\mathbb{R}^n$ the vector whose components are all equal to $1$.}

Let us assume we have a vocabulary of $M$ words $V=\{w^1, \dots, w^M\}$. To each word $w_i$ we will assign a vector  $x^i \in  \mathbb{R}^n$, where $x$ is called a word vector, or word embedding. We will call  $\mathcal{W} = \{x^1, \dots x^M \}$ the \emph{embedding space} and  $n$ is called the dimension of the embedding space.  
We will denote the word vector associated to a word w by $\overrightarrow{w}$

We define a distance function in the embedding space as any function $d$ that satisfies:
\begin{enumerate}  
   \item $d:\mathcal{W}\times \mathcal{W}
   \rightarrow \mathbb{R}$. 
   \item $d(x,x)=0$
\end{enumerate}  
In particular, in this paper, we use the cosine similarity metric:
\begin{equation}\label{eq:int.1}
  d(x,y) = 1 - 
  \frac{ \inner{x}{y}}{ \norm{x}\norm{y}}.
\end{equation}
Note that,  for all $x, y\neq 0$, $d(x,y)$ is  non-negative  and the following identity holds
\begin{equation}\label{eq:int.2}
d(x,y) = \frac 1 2 \norm{\frac{\textstyle x}{  \textstyle \norm{x}}-\frac{\textstyle y}{  \textstyle \norm{y}}}^2
\end{equation}
In particular,  $d(x,y)=0$ for $x, y\neq 0$ imply that $x/\norm{x}=y/\norm{y}$.

\section{Embedding-creation problems} \label{sec:EMB_CREATION} 

 Most commonly, word embedding spaces are learned from a corpus and once learned new vectors are not added. Word embeddings can be combined for creating sentence representations but it is more unusual to add new vector representations. In this section, we discuss the typical problem of word retrieval given a word vector, and introduce the problem of how to create new "melting-pot" vectors that could be added to the embedding space. 

  \subsection{Word-retrieval problem}
  
  In a number of NLP problems, a question to solve is to determine, given a vocabulary $V$ and a vector $x \in \mathbb{R}^n$, which is the word in $V$ that would correspond to this vector. A common solution to this problem, is to find the vector $y \in W$ such as $d(x,y)$ is minimized. We call this problem as the \emph{word finding problem}, and formally define it as a minimization problem. 

  \begin{equation}
     v^* = \underset{v \in V}{\min}\, d(x,\overrightarrow{v}) 
  \end{equation}    

 We can also define a partition of $\mathbb{R}^n$ in $M$ classes $\{c_1, \dots, c_m \}$ according to $W$. We say that $x \in c_i$ if $v^i = min_{v \in V} d(x,\overrightarrow{v})$.

\subsection{Melting-pot embedding problem}\label{ssu.m.pot}

We envision one scenario different to the previous one in which, instead of finding a vector associated to a word, we would like to create an embedding that acts as a  ``melting pot'' of the meanings for two different words. This is, it will capture some semantic aspects of the two words. In practice, we assume that such fusion of significations can be achieved by constructing a vector representation that is ``as close as possible'' to the embeddings of both original words. The underlying assumption is that, if such embedding had an associated word, that word would have  some semantic relationships with \emph{both} original words. In fact, in many applications we do not require the existence of such a word, it is sufficient with representing that ``wordless meaning'' with an embedding. 

Given two words  $v^1$ and $v^2$ in the vocabulary $V$, with associated embedding space $W$, we define the embedding-creation problem as the problem of finding a vector $x \in \mathbb{R}^n$ such that $d(x,\overrightarrow{v^1})=d(x,\overrightarrow{v^2})$ and $d(x,\overrightarrow{v^1})$ is minimal.

We will organize the analysis of this problem and its solution by addressing the following questions:

\begin{enumerate}
 \item Under which conditions such a vector exists?
 \item Which mathematical approach should be used to solve the problem?
\end{enumerate}

{We will  leave the  answer of these questions for the next subsection in the context of a generalization of the embedding problem.}

\subsection{Generalizations of the embedding creation problem}


{In this Subsection we consider different ways to generalize the problem of creating an embedding  from the composition of two different word embeddings. In particular, we consider:}
 
 \begin{itemize}
 \item Problem 1
 \begin{itemize}
 \item Is it possible to prove the existence of an $n$-dimensional vector $x$ which is at the same distance from $N$ other vectors? 
 \item How to find such vectors?
 \item{Among those equidistant vectors, is there one vector closer to the initial given vectors?}
 \end{itemize}
 \item Problem 2
 \begin{itemize}
 \item Under which conditions it is possible to find an $n$-dimensional vector $x$ which is at given distances from $N$ other vectors? 
 \item How to find such vectors?
 \end{itemize}
 \end{itemize}

{Note that solving Problem 1 above provides the answers for the questions posed in the Subsection \ref{sec:EMB_CREATION}.}

\subsubsection{Word-vector equidistant from $N$ other vectors} \label{ssec:EQUIDISTANT_N}

We will start considering Problem 1 above. 
Let us assume that we have a set of $N$ non-zero vectors $\{v_1,\ldots,v_N\}$, 
{whose normalized vectors are not all equal}.
The problem we are interested in is to find $x\in \R^n-\{0\}$ such that
\[
d(x,v^N)=d(x,v^j)\quad j=1,\ldots,N-1.
\]
Note that this is equivalent to find $x\in \R^n-\{0\}$ such that
\[
\left<\frac{x}{|x|},\frac{\overrightarrow{v^N}}{|\overrightarrow{v^N}|}-\frac{\overrightarrow{v^j}}{|\overrightarrow{v^j}|}\right> =0,\quad j=1,\ldots,N-1,
\]
that is, defining vectors
$w_i=\overrightarrow{v^N}/|\overrightarrow{v^N}|-\overrightarrow{v^j}/|\overrightarrow{v^j}|$ $j=1,\ldots,N-1$ and denoting the linear span of these vectors by $S_1$, the problem is to find a non-zero vector $x\in S_1^\bot$, the orthogonal complement of $S_1$. 
Let $m$ denote the dimension of $S_1$. 
{Note that $m\leq \min\left\{N-1,n\right\}$, and $S_1^\bot =\{0\}$ if, and only if, $m=n$, which provides a necessary and sufficient condition for our problem of interest to have a solution.
To describe all solutions of this problem, we could compute an orthogonal basis of $S_1^\bot$.
The previous discussion  clearly answers the first two questions posed in Problem 1.
To answer the last  question, we will assume that $S_1^\bot \neq\{0\}$ in order to ensure that the problem has a solution. 
Note that the third question of Problem 1 amounts to solve the following optimization problem}
\begin{equation}\label{emb.propb1}
\begin{aligned}
\min_{x} \quad & d(x,\overrightarrow{v^N})\\ 
\textrm{s.t.} \quad & d(x,\overrightarrow{v^N}) = d(x,\overrightarrow{v^j}) \quad j=1,\ldots,N-1,
\end{aligned}
\end{equation}
or, by trivial calculations,
\begin{equation}\label{emb.propb2} 
\begin{aligned}
\min_{x} \quad & d(x,\overrightarrow{v^N})\\ 
\textrm{s.t.} \quad & \left< \frac{x}{|x|},\frac{\overrightarrow{v^N}}{|\overrightarrow{v^N}|}-\frac{\overrightarrow{v^j}}{|\overrightarrow{v^j}|}\right> =0 \quad j=1,\ldots,N-1.
\end{aligned} 
\end{equation} 

 The above problem can be reformulated as follows:
\begin{equation} 
\begin{aligned}
\min_{\hat x} \quad & d\left(\hat x,\frac{\overrightarrow{v^N}}{|\overrightarrow{v^N}|}\right)\\
\label{eq:EXIST_B1}
\textrm{s.t.} \quad & \hat x\in S
\end{aligned}
\end{equation}
where $S=\left\{x\in \R^n\,:\,|x|=1\right\}\cap S_1^\bot$ and $S_1$ is the linear space defined above.
Note that in \eqref{eq:EXIST_B1} we are minimizing a continuous function over the compact set $S$, therefore the above problem has at least one solution.

 \begin{proposition}\label{prop:opt.refor} 
 Problem \eqref{eq:EXIST_B1} has a unique solution given by
 \begin{equation}
 x^\ast=\frac{ P_{S_1^\bot}\left(\overrightarrow{v^N}/|\overrightarrow{v^N}|\right)}{ \left|P_{S_1^\bot}\left(\overrightarrow{v^N}/|\overrightarrow{v^N}|\right)\right|}
 \label{eq:OPT_FINAL}
 \end{equation}
 \end{proposition}
We call the vector in \eqref{eq:OPT_FINAL} the {\it Optimal Synthesis Embedding (OSE)} of  vectors $v_1$,..,$v_N$.

\begin{remark}\label{rm:prop.1a}
Since it holds $d(\alpha x,\beta y)=d(x,y)$ $ \forall x, y\in \R^n-\{0\}$ and $\alpha,\beta>0$, it is easy to see that the set of solutions of Problem \eqref{emb.propb1} is the one-dimensional linear space $S=\{tx^\ast\,|\,t\in\mathbb{R}, t>0\}$ where $x^\ast$ is given by \eqref{eq:OPT_FINAL}.
\end{remark}

\begin{remark}\label{rm:prop.1}
Note that in the case of two vectors, i.e. $N=2$, is easy to obtain the following explicit expression for the solution in \eqref{eq:OPT_FINAL}, $x^\ast =(\hat v^1+\hat v^2)/\left\|\hat v^1+\hat v^2\right\|$ where $\hat v^1=v^1/\|v^1\|$ and $\hat v^2=v^2/\|v^2\|$. Thus, in this case, the solution set $S$ of the Problem \eqref{emb.propb1} includes the arithmetic mean of the normalized vectors $ x=(1/2)(\hat v^1+\hat v^2)$.
For $N>2$, the arithmetic mean of the normalized vectors $\hat v=(1/N)\left(\sum_{i=1}^N \hat v^i\right)$ where $\hat v^i=v^i/\|v^i\|$ for $i=1,\ldots,N$ does not necessarily belong to the solution set of Problem  \eqref{emb.propb1}. Note that for $N=3$, taking $v^1=(1,0,0)$, $v^2=(0,1,0)$ and $v^3=(1,1,1)$, simple calculations show that
\[
\inner{\frac{\te 1}{\te 3}\left( \sum_{i=1}^3 \hat v^i\right)}{\hat v^3-\hat v^1}
=\frac{ \textstyle 1}{  3}+\frac{ \textstyle 1}{  3\sqrt{3}}
\]
which means that $x=(1/3)\left(\sum_{i=1}^3 \hat v^i\right)$ does not satisfy the constraints in \eqref{emb.propb2}.
\end{remark}

\subsubsection{Word-vector at given distances of $N$ other vectors}  \label{sec:ATGIVENDISTANCE_N}

Next we deal with Problem 2. Let us assume initially that we have a set
of l.i. vectors $B=\{v_1,\ldots,v_N\}\subset \R^n$ and a vector of 
scalars $\alpha=(\alpha_1,\ldots,\alpha_N)\neq 0$   with $N<n$.
We need to find $x\in \R^n$ such that $d(x,v_i)=\alpha_i$,  $i=1,\ldots,n$.
The definition of the cosine similarity metric yields the following reformulation of  this problem: 
\begin{equation}
\label{prob2.refor}
\mbox{find } x\in \R^n \mbox{ with } \|x\|=1 \mbox{ 
and such that } Vx=w
\end{equation}
where $V$ is the matrix whose  row vectors are the vectors of $B$ after a normalization and
$w=(1-\alpha_1,\ldots, 1-\alpha_N)^t$.

Under the above assumptions, the linear system in \eqref{prob2.refor}, $Vx=w$, has solutions.
Given any particular solution $x_0$ of this system, the next result shows a necessary and sufficient for problem \eqref{prob2.refor} to have a solution.

\begin{proposition}\label{prop:P2.related.refor}
 Let $x_0$ a solution of the linear system in \eqref{prob2.refor}. The problem \eqref{prob2.refor} has a solution if, and only if,
\begin{equation}
\label{eq:nec.suf.cond}
1\geq \norm{x_0-P_{S_0}(x_0)}
=\norm{V^t\left(VV^t\right)^{-1}w}
\end{equation}
where $S_0=\left\{\hat x\in \R^n\, |\, V\hat x=0\right\}\neq\{0\}$. If the above condition holds, then $ {x^\ast = x_0-P_{S_0}(x_0)+t^\ast x_1}$, with $0\neq x_1\in S_0$ and
$t^\ast=  \sqrt{\left(1-\left\|V^t\left(VV^t\right)^{-1}w
\right\|^2\right)/\norm{x_1}^2}$, is a solution of problem \eqref{prob2.refor}.
\end{proposition}

\begin{remark}\label{rm:prop.2}
In particular, if \eqref{eq:nec.suf.cond} holds in equality, then $x^\ast=V^t\left(VV^t\right)^{-1}w$ is a solution.
In adddition, if $N=n$, then $S_0= \{0\}$ and vector $x^\ast=V^{-1}w$ is  the unique
solution of the linear system $Vx=w$. 
Hence, in this case, Problem 2 has solution if, and only if, $\left\|V^{-1}w\right\|=1$.
Finally, note that in the case where the set  $B=\{v_1,\ldots,v_N\}$ is l.d.
we can use, for example, Gaussian elimination  
to check if the linear system $Vx=w$  has solutions, and, if this is the case, reduce the analysis to the case presented above.
\end{remark}

\begin{remark}\label{rm:decrea.dist}
The particular case in which $w=(1-\alpha){\bf e}$, where $\alpha\in [0,2]$, corresponds to the solution of  Problem 2 being at the same distance $\alpha $ from all the vectors in $B$. In this case, condition \eqref{eq:nec.suf.cond} becomes $|1-\alpha|\norm{V^t\left(VV^t\right)^{-1}e}\leq 1$, whose solution is $\alpha\in[\underline{\alpha},\overline{\alpha}]$ where
$\underline{\alpha}=\max\left\{0, 1-1/\norm{V^t\left(VV^t\right)^{-1}e}\right\} $ and $\overline{\alpha}= \min\left\{2,1+1/\norm{V^t\left(VV^t\right)^{-1}e}\right\} $.
Hence, if $\norm{V^t\left(VV^t\right)^{-1}e}\leq 1$, then we could take $\alpha=\underline{\alpha}=0$.
This means, in view of the comments after equation \eqref{eq:int.2}, that all the vectors in $B$ are equal after normalization, which contradicts our linear independence assumption. If $\norm{V^t\left(VV^t\right)^{-1}e}> 1$ then, among the  equidistant to the vectors in $B$, the closest vector to the vectors in this set is at a distance of $\underline{\alpha}=1-1/\norm{V^t\left(VV^t\right)^{-1}e}$.
Note that in Subsection~\ref{ssec:EQUIDISTANT_N}, when solving Problem 1, finding this value was not required. 
In addition, considering $\alpha=\underline{\alpha}$ and  in view of the discussion after \eqref{eq:nec.suf.cond}, it follows that 
$ x=(1-\underline{\alpha})V^t(VV^t)^{-1}e = (1/\left\|V^t(VV^t)^{-1}e\right\|)V^t(VV^t)^{-1}e$
is a solution of Problem 1, which provides an alternative to the expression  in  \eqref{eq:OPT_FINAL}.
Now, let us assume that $\underline{\alpha}\neq 0$. 
Take $t\in\mathbb{R}$ and $u=(u_1,\ldots,u_N)\in\mathbb{R}^N$ such that $u_i<u_{i+1}$ for $i=1,\ldots,N-1$ and $V^t\left(VV^t\right)^{-1}u\neq 0$. 
Consider  the problem of finding a vector $x\neq 0$ such that $d(x,v_i)=\underline{\alpha}+t u_i$ for $1,\ldots,N$. 
Note that, in view of the previous discussion, for $t=0$ this problem has a solution. 
Using \eqref{eq:nec.suf.cond}, it follows that this problem has solution for all $t$ such that $\norm{V^t\left(VV^t\right)^{-1}((1-\underline{\alpha}) e-tu)}\leq 1$. 
Simple calculations show that this is equivalent to the relation
$at^2+bt+c\leq 1$ where $a=\norm{V^t\left(VV^t\right)^{-1}u}^2$, 
$b=2(1-\underline{\alpha})\inner{u}{\left(VV^t\right)^{-1}e}=2\inner{u}{\left(VV^t\right)^{-1}e}/\norm{V^t\left(VV^t\right)^{-1}e}$ and 
 $c=(1-\underline{\alpha})^2\norm{V^t\left(VV^t\right)^{-1}e}^2=1$.
 Thus, it follows that the  problem described above has solution for any $t\in[-b/a,0]$ if $b\geq 0$, and $t\in [0,-b/a]$ otherwise.
\end{remark}

\section{Evaluation of the Optimal Synthesis Embeddings} \label{sec:OSE_APP}


In this paper we propose the Evaluation of the Optimal Synthesis Embeddings method (also denoted by OSE) in two different ways: 1) By computing a unique representation of multiple words that belong to the same category; and 2) By computing an OSE for a sentence, which is computed from the set of embeddings corresponding to each word in the sentence. 


\subsection{Data augmentation}    

     There are classification problems in which the number of available examples from some of the classes is reduced. This lack of data affects the performance of machine learning methods. Let us assume a word classification problem where each word is represented using a given word-embedding representation.    We propose the use of OSE as a way to generate new examples from a given class. The method, that we call OSE-Augmentation will take $k$ random words of the same class from the train set, and construct the OSE representation, which is then added as a new example of the class. The rationale behind the use of OSE for data augmentation is that a word vector representation that is a minimum distance of other $k$ instances that belong to the class, is a good exemplar of the class. In addition, being equidistant, it is not biased by any member of the class. 
          
     As an alternative way to create augmented examples, we use the Bag of Vectors (BOV) method,  where the arithmetic mean of word-vectors for all words in the sentence is computed.  The word generation process for data augmentation can be  repeated $K$ times, guaranteeing that the same set of seeded words will not be used to create the augmented example. $k$ and $K$ are parameters of the method, the first, specifies the number of \emph{seeded} words, the second, the number of examples that are added to training set of the given class. 
  
   \subsection{Sentence representation}    

     An ubiquitous problem in NLP is how to create universal sentence representations that could be transferred among NLP tasks. There are two main approaches to sentence representation. One possibility is to combine the word embedding representation of all words in the sentence. The other approach is to directly generate a sentence representation. Notice that, when combining word embeddings, these could correspond to fixed (pre-computed) embedding or generated from the sentence by models that use a contextual representation. 

    Computing OSE to represent sentences can be seen  as an alternative to the  BOV method which is extensively used for this purpose. However, it is important to remark that assuming that the sentence embedding should be at the same distance of all the elements in the sentence, as OSE does,  is indeed a strong assumption. We can expect that in many realistic situations not all the elements in the sentence have the same semantic relevance.

    Furthermore, assuming a fixed embedding representation, OSE will clearly be ineffective to distinguish between a grammatically correct sentence and any random permutation of the words in the sentence.  However, the BOV representation, that has been extensively applied as a computational fast algorithm to create sentence embeddings, suffers from the same limitation.

  \section{Experiments}  \label{sec:EXPE}
  
  In this section we evaluate the performance of the OSE representation and compare it with a number of state of art methods for sentence embeddings. We introduce first the experimental framework, and then present and discuss the results of the experiments. 
   
\subsection{Experimental framework}

  \subsubsection{Data augmentation problems}   
  
   For the data augmentation analysis, we use the SEMCAT dataset introduced in \cite{Senel_et_al:2018}. It contains more than $6500$ words semantically grouped under $110$ categories.  We removed all possible words repetitions in each category, and keep only those categories with $100$ or more words. 


   After filtering, $30$ categories remained. We then use data augmentation to solve the $30$-class classification problem of predicting the category of a word given its word-embedding. Each set of words was divided into two equally-sized  train and test sets. From the train set, we selected subsets of words of size $k$, for $k \in \{ 2,5,10,15,20 \}$. These subsets were employed to create 
   augmented word vectors using the \emph{OSE} and \emph{BOV} strategies and three different types of word embeddings with dimension $n=300$: fasttext \cite{Joulin_et_al:2016},  glove  \cite{Pennington_et_al:2014}, and word2vec   \cite{Mikolov_et_al:2013a}. All in all, we have $5 \time 2 \times 3 = 30$ possible configurations of the classification experiment.

  \subsubsection{Sentence embedding problems}

   For sentence embedding evaluation, we use the \emph{SentEval} library \cite{Conneau_and_Kiela:2018}  that comprises a number of downstream tasks\footnote{Available from \url{https://github.com/facebookresearch/SentEval}}. We have used  $14$ of the original $18$ transfer tasks and $10$ recently added probing  tasks \cite{Conneau_et_al:2018}. The NLI, STSBenchmark, SICKRelatedness,  and COCO tasks were not considered due to the computational cost associated to processing the large datasets.

\begin{table}[htbp]
\begin{center} 
\begin{tabular}{r|r|rrrr}\hline

Emb./Class. &Method  &  KNN & LDA & NC & RC \\ \hline
fasttext& no augm.  &        0.598 & 0.633 & 0.636 & 0.630 \\
        &   BOV     &        0.620 & 0.628 & {\bf{0.637}} & 0.632 \\
        &   OSE     &        0.617 & 0.627 & {\bf{0.637}} & 0.627 \\ \hline
glove   &   no augm. &        0.568 & 0.629 & 0.602 & {\bf{0.633}} \\
        &   BOV      &        0.602 & 0.627 & 0.602 & 0.630 \\
        &   OSE      &        0.592 & 0.627 & 0.600 & 0.632 \\  \hline
word2vec & no augm. &        0.576 & 0.601 & 0.605 & 0.599 \\
         &  BOV     &        0.591 & 0.598 & 0.608 & 0.602 \\
         &   OSE    &        0.584 & 0.588 & {\bf{0.613}} & 0.598 \\ \hline
\end{tabular}
 \end{center}
    \caption{Results of four of the best classifiers for the category classification problem, with and without data augmentation. KNN: K-nearest neighbors classifier; LDA: Linear Discriminant Analysis; NC: Nearest Centroid classifier, and Ridge regression. Best results for each type of embedding are shown in bold.}
    \label{tab:DA_Comparison}
\end{table}

\begin{table*}[]
   \caption{Results of the multi-layer perceptron for probing tasks using different embedding methods.}
   \label{tab:Content_Results_NN}
   \vskip 0.15in
  \tiny
  \centering
  \begin{sc}
    \begin{tabular}{c|rr|rr|rr|rr|rr|r|rr}
\toprule
     Embedding&  \multicolumn{2}{c|}{fasttext}  &  \multicolumn{2}{c|}{glove} &  \multicolumn{2}{c|}{word2vec} &   \multicolumn{2}{c|}{BERT} &  \multicolumn{2}{c|}{RoBERTa} &    USE & \multicolumn{2}{c}{total} \\ \hline
Benchmark&       BOV &       OSE &    BOV &    OSE &      BOV &      OSE &    BOV &    OSE &      BOV &      OSE &  &      BOV &      OSE   \\ \hline \hline
Length & 36.83  & 50.70  & 35.25  & 40.23  & 35.12  & 49.80  & 70.11  & 66.87  & {\bf{71.60}}  & 59.14  & 65.46   &2 &3 \\
WordContent & 80.55  & 86.91  & 76.83  & 77.52  & 36.32  & {\bf{86.98}}  & 42.92  & 36.28 & 20.85 & 4.81 & 70.15  &2 &3 \\
Depth & 27.94  & 28.20  & 27.29  & 25.39  & 27.01  & 26.65  & 32.71  & {\bf{33.49}}  & 24.02  & 24.68  & 27.70    &2 &3 \\
TopConstituents & 61.37 & 65.18 & 57.45 & 60.43 & 56.78 & 57.70 & 66.56 &{\bf{73.75}} & 33.85  & 29.55  & 62.82   &1 &4 \\
BigramShift & 50.22 & 49.91 & 50.15 & 49.92 & 49.93 & 50.16  & 82.86  & {\bf{83.46}}  & 50.88  & 51.47  & 60.48   &2 &3 \\
Tense & 86.61  & 86.50  & 83.00  & 84.93  & 84.29  & 85.72  & 87.45  & {\bf{87.72}}  & 59.64  & 58.73  & 80.12    &2 &3 \\
SubjNumber & 79.74 & 80.64 & 77.98 & 79.13 & 79.79  & 81.12  & {\bf{85.01}}  & 83.87  & 63.81  & 62.75  & 74.61   &2 &3 \\
ObjNumber & 78.84 & 78.70 & 75.75 & 75.29 & 77.81 & 79.17 &{\bf{80.95}} & 80.68  & 62.62  & 62.36  & 72.79        &4 &1 \\
OddManOut & 50.15  & 52.97 & 50.89 & 52.49  & 49.87 & 51.80 & 60.98 &{\bf{63.61}} &49.78  & 49.81  & 54.36        &0 &5 \\
CoordinationInversion &52.23 & 52.27 & 53.23 & 52.74 &53.26 &51.95 &68.34 &{\bf{68.52}} &52.67 &52.62 &54.54      &3 &2 \\ \hline
Total &3   &7      &4  &6     &2  &8    &4    &6   &7   &3  &     &20  &30  \\ 
\bottomrule
\end{tabular}    
  \end{sc}
  \vskip -0.1in
\end{table*}

   For sentence embeddings, we investigate both, context-independent and context-dependent word embeddings. As exemplars of the first class we consider fasttext \cite{Joulin_et_al:2016},  glove \cite{Pennington_et_al:2014}, and word2vec \cite{Mikolov_et_al:2013a}. Context-dependent embeddings comprise BERT \cite{Devlin_et_al:2018} and RoBERTa \cite{Liu_et_al:2019a}, as implemented by the HuggingFace library \cite{Wolf_et_al:2019}. In addition, we compare with the Universal Sentence Encoder (USE)  representation \cite{Cer_et_al:2018}. When dealing with context-dependent encodings such as BERT, the number of vectors for a given sentence can be greater that the number of words in the sentence. In this case, we compute the BOV and OSE sentence representations using the vectors from the resulting tokens.  
      
   As another dimension of analysis, we consider two different classification approaches: The multi-layer perceptron, as implemented in the SentEval repository \cite{Conneau_and_Kiela:2018}, and an KNN method based on the cosine distance. This latter classifier is included since class assignment in KNN is based on the computation of the distances to the nearest neighbors. Therefore, we would like to determine whether a strategy such as OSE, that minimizes the distance between the sentence embedding and the embeddings of the words comprised by sentence, is particularly beneficial for KNN.  

\subsection{Results for data augmentation}

\begin{figure}[htbp]
    \begin{center}
   \includegraphics[width=8.5cm]{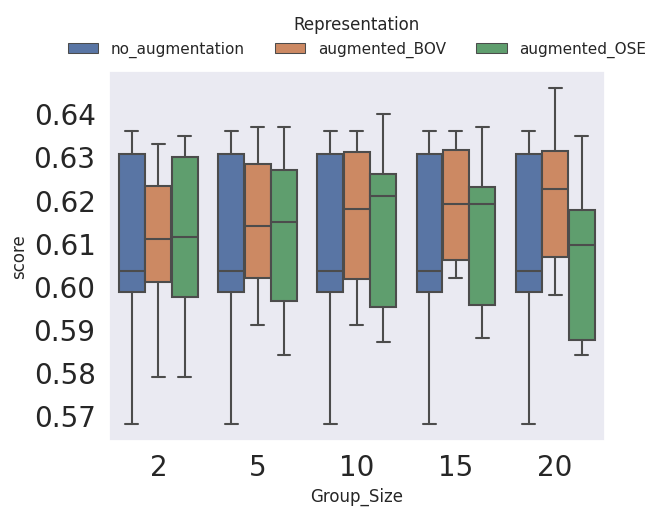}     
   \caption{Comparison between OSE and BOV for the data augmentation problem. }
  \label{fig:DA_Results}
 \end{center}
  \end{figure}

 We applied $17$ classifiers to solve the word category classification problem with and without data augmentation. The results for  four of the  classifiers that produced the higher accuracy, and group size ($k=5$) are shown in Table~\ref{tab:DA_Comparison},  where the best results for each type of embedding are highlighted in bold. Figure~\ref{fig:DA_Results}, for each group size, the accuracies when considering all classifiers and the three types of embeddings.

\begin{table*}[]
    \caption{Results of Multi-layer perceptron for sentence classification and natural language inference tasks. 
}
    \label{tab:Semantic_Results_NN}
    \vskip 0.15in
 \tiny
 \centering
    \begin{sc}
    \begin{tabular}{c|rr|rr|rr|rr|rr|r|rr}
\toprule
     Embedding&  \multicolumn{2}{c|}{fasttext}  &  \multicolumn{2}{c|}{glove} &  \multicolumn{2}{c|}{word2vec} &   \multicolumn{2}{c|}{BERT} &  \multicolumn{2}{c|}{RoBERTa} &    USE & \multicolumn{2}{c}{total} \\ \hline
Benchmark&       BOV &       OSE &    BOV &    OSE &      BOV &      OSE &    BOV &    OSE  &      BOV &      OSE &  &      BOV &      OSE     \\ \hline \hline
MR & 77.47  & 77.42  & 76.65  & 74.07  & 76.37  & 76.15  & {\bf{79.36}}  & 78.16  & 57.14  & 57.44  & 75.30 & 4& 1\\
CR & 79.39  & 79.65  & 79.73  & 76.98  & 77.75  & 77.72  & {\bf{84.72}}  & 83.05  & 66.07  & 63.95  & 81.45 & 4& 1 \\
SUBJ & 91.76  & 91.64  & 90.66  & 88.44  & 90.07  & 90.07  & 94.10  & {\bf{94.23}}  & 74.15  & 68.72& 91.75 & 3& 1\\
MPQA & 87.81  & 87.84  & 87.47  & 87.25  & {\bf{88.13}}  & 87.86  & 79.10  & 79.87  & 69.58  & 69.36& 87.29 & 3& 2\\
TREC & 84.60  & 84.80  & 83.80  & 82.20  & 83.40  & 84.40  & 89.60  & 90.00  & 60.00  & 52.60 & {\bf{92.40}}& 2& 3\\
SST2 & 82.32  & 81.00  & 80.72  & 77.48  & 79.79  & 79.57  & {\bf{85.17}}  & 82.48  & 59.69  & 58.54 & 81.16& 5& 0\\
SST5 & 44.71  & 43.30  & 43.53  & 40.27  & 42.94  & 42.40  & {\bf{45.43}}  & 44.12  & 29.73  & 28.28 & 43.44& 5& 0\\
MRPC & 71.83  & 70.78  & 68.81  & 68.00  & 72.41  & 72.23  & {\bf{73.80}}  & 72.46  & 67.54  & 64.23 & 72.81& 5& 0\\
SICKEntailment & 79.28  & 77.59  & 78.87  & 74.47  & 75.91  & 77.63  & 76.66  & 74.00 & 64.54  & 63.57  & {\bf{82.32}} &4 &1 \\ \hline
Total & 6  &  3    &9  & 0    & 6 & 2   & 6   & 3  & 8  & 1 &     & 35 & 9\\
\bottomrule
    \end{tabular}
    \end{sc}
    \vskip -0.1in
\end{table*}

 \begin{table*}[htbp]
     \caption{Results of the neural network classifier for semantic textual similarity tasks.}
     \label{tab:STS_Results}
     \vskip 0.15in
      \tiny
      \centering
      \begin{sc}
    \begin{tabular}{c|rr|rr|rr|rr|rr|r|rr}
\toprule
     Embedding&  \multicolumn{2}{c|}{fasttext}  &  \multicolumn{2}{c|}{glove} &  \multicolumn{2}{c|}{word2vec} &   \multicolumn{2}{c|}{BERT} &  \multicolumn{2}{c|}{RoBERTa} &    USE & \multicolumn{2}{c}{total}  \\ \hline
Benchmark&       BOV &       OSE &    BOV &    OSE &      BOV &      OSE &    BOV &    OSE &      BOV &      OSE & &      BOV &      OSE    \\ \hline \hline
STS12 & 0.607  & 0.585  & 0.567  & 0.550  & 0.528  & 0.536  & 0.480  & 0.499  & 0.371  & 0.337  & {\bf{0.683}}  &3 &2 \\
STS13 & 0.661 & 0.594  & 0.603  & 0.574  & 0.560  & 0.606  & 0.588  & 0.564  & 0.342  & 0.247  & {\bf{0.719}}  &4 &1 \\
STS14 & 0.672  & 0.652  & 0.608  & 0.565  & 0.614  & 0.638  & 0.577  & 0.577  & 0.391  & 0.315  & {\bf{0.736}}  &3 &1 \\
STS15 & 0.716  & 0.732  & 0.641  & 0.629  & 0.630  & 0.712  & 0.634  & 0.660  & 0.396  & 0.327  & {\bf{0.813}}  &2 &3 \\
STS16 & 0.679  & 0.723  & 0.582  & 0.649  & 0.547  & 0.696  & 0.621  & 0.648  & 0.325  & 0.294  & {\bf{0.788}}  &1 &4 \\ \hline
Total & 3      & 2      & 4      & 1      &0       & 5      &1       &3       & 5      &0       &  &13  & 11          \\ 
\bottomrule
    \end{tabular}
      \end{sc}
      \vskip -0.1in
\end{table*}

  The analysis of Table~\ref{tab:DA_Comparison} reveals that data augmentation improves the classification performance for the KNN classifier and to a lesser extent for the Nearests Centroid classifier. Results between BOV and OSE are similar, although for the word2vec embedding OSE obtains the highest accuracy.

  When computing the mean of the classifiers accuracies, it is clear that data augmentation improves the baseline classifier. Results for BOV and OSE are similar, although when $n=20$ there  is a noticeable decrease in the accuracy of OSE.

.
  
\subsection{Results for sentence embedding problems}

 \subsubsection{Probing benchmark}
 
 The results of the comparison between the BOV and OSE methods for the probing tasks  benchmark are shown in Table~\ref{tab:Content_Results_NN} where the best absolute results among all embedding methods and sentence embedding strategy is emphasized in bold. The last column and last row of each table indicates the number of times that BOV has outperformed OSE and vice-versa. 

 Table~\ref{tab:Content_Results_NN} shows the results achieved by the multi-layer perceptron classifier. It can be seen in the table that for seven of the ten problems the best results are achieved using the OSE representation. For two problems, WordContent and TopConstituents, the improvement of the best OSE result over the best BOV result  was of over a $6\%$. In global terms, OSE achieves better results than BOV in $30$ of the $20$ combinations of embedding methods and problems. Among the fixed embeddings, it is particularly effective for word2vec.

Table~\ref{tab:Content_Results_KNN} (Anex B) shows the results of the KNN classifier for all the probing tasks. For almost all combination of embeddings and problems, KNN is outperformed by the classification accuracy of the multi-layer perceptron. However, the KNN results help us to investigate the relative performance of OSE with respect to BOV in a different classification scenario where class assignment depends on the explicit computation of the distances between sentence embeddings. In can be appreciated in the table that for $7$ of the $10$ problems OSE outperforms BOV. When KNN is used, OSE outperforms BOV in $36$ of the $50$ scenarios of comparison. 

All in all, the analysis of the tables shows that OSE can contribute to improve the best classification results for $7$ of the problems, and clearly outperforms BOV in most of scenarios considered. Remarkably, OSE can improve the results of fixed embeddings such as word2vec, and also improves the results for BERT, which is the absolute best performing embedding algorithm. BERT produces the best absolute results for $8$ of the $10$ problems. In $6$ of these cases,  the results are achieved by using OSE.

It should be noted that some of the tasks included in the probing dataset cannot be solved by a combination of fixed word embeddings and sentence embeddings such as BOV and OSE that are invariant to permutation of the words order. For these problems, our results show that the best strategy is the use of OSE together with BERT embeddings.

\subsubsection{Classification, Natural Language Inference, and Semantic Similarity tasks}

We also investigated the performance of OSE for benchmarks corresponding to other types of NLP tasks. Tables~\ref{tab:Semantic_Results_NN} and~\ref{tab:Semantic_Results_KNN} (Annex B) show the results of OSE and BOV for the evaluated benchmarks. In the tables, the first $7$ rows correspond to classification tasks. MRPC is a paraphrase detection task and SICK-Entailment is an example of natural language inference task.

The results of the multi-layer perceptron model are shown in Table~\ref{tab:Semantic_Results_NN}. The analysis of the table reveals that, in contrast to the probing tasks, for these tasks, BOV overwhelmingly outperforms OSE. For only one of the $9$ problems, OSE produces the best absolute results.  BOV outperforms OSE for $35$ of the $45$ benchmarks.   

Table~\ref{tab:Semantic_Results_KNN} shows the results when KNN is applied. Also in this case, OSE produces the best absolute results only in one of the $9$ problems. However, both algorithms have similar results when considering all combinations of benchmarks and embeddings. OSE slightly edges BOV by outperforming it for $24$ of the $45$ benchmarks.

\subsubsection{Semantic textual similarity benchmark}

 Finally, we also investigated OSE for the semantic textual similarity benchmark that includes five problems. For these problems, the quality of the sentence embedding is measured in terms of the correlation  between the cosine-distance between two sentences and a human similarity score. This means that  the task is unsupervised and no classification method is involved. 
  
 The results for the Semantic textual similarity benchmark are shown in Table~\ref{tab:STS_Results}. The analysis of the table shows that USE produces the best absolute results for the five tasks. After USE, the best embedding results are achieved using fasttext for each benchmark. For three of the five cases, BOV outperforms, and for the remaining two OSE wins. 


\section{Discussion}  \label{sec:CONCLU}

 In this work, a new approach to create embedding representation of multiple words by the composition of embeddings has been introduced. A key element of our proposal is enforcing that the representation should be at the same distance of all the word embeddings for the constituent words. We have also derived the solution to the problem of creating a vector representation at given distances of a set of vectors.  We have empirically evaluated the approach and concluded that it can outperform BOV on probing tasks. However, for other sentence classification tasks the method does not show any clear advantage over BOV. 


A possible limitation of our approach is that it can be sensitive to outliers. The rational behind the equidistant representation assumes that all words are equally relevant. The presence of an outlier in the set of words can produce that the distance from the new vector representation to all the other embeddings can be very long. 



\section*{Acknowledgements}

R. Santana acknowledges partial support by the Research Groups 2022-
2024 (IT1504-22) and the Elkartek Program (KK-2020/00049, KK-2022/00106,
SIGZE, K-2021/00065) from the Basque Government, and the PID2019-104966GB-
I00 and PID2022-137442NB-I00 research projects from the Spanish Ministry of
Science.


\bibliographystyle{icml2024}
\bibliography{wordless_emb_references}

\newpage
\appendix
\onecolumn
\section{Proofs}



Proof of Proposition \ref{prop:opt.refor}
\begin{proof}
 Let $B=\left\{w_1,\ldots,w_{k}\right\}$, with $k\leq n$, be an orthonormal vector
 basis of $S_1^\bot$.
Then for any vector $x\in S_1^\bot$ 
 there exists a (unique) set of real scalars $\alpha_j$ $j=1,\ldots,k$ (the coordinates of $x$ in the base
 $B$) such that $x=\sum_{j=1}^{k}\alpha_j w_j$ and, in addition, it holds $|x|^2=\sum_{{j}=1}^{k}\alpha_{{j}}^2$. Then, we can reformulate Problem \eqref{eq:EXIST_B1} as follows:
\begin{equation}
\begin{aligned}
 \min_{(\alpha_1,\ldots,\alpha_{k})\in R^{k}}\quad &
 p (\alpha_1,\ldots,\alpha_{k}) = 1-\sum_{j=1}^{k}\alpha_j\beta_{j}\\ 
\textrm{s.t.} \quad & q (\alpha_1,\ldots,\alpha_{k})=\sum_{j=1}^{k}\alpha_j^2=1
\label{eq:REFORM} \\
\end{aligned}
\end{equation}
 where $\beta_{j} = \left<w_j,\overrightarrow{v^N}/|\overrightarrow{v^N}|\right>$ for $j=1,\ldots,k$.
 Note that $\sum_{j=1}^{k}\beta_{j\,i}w_j =
 P_{S_1^\bot}\left(\overrightarrow{v^N}/|\overrightarrow{v^N}|\right)
 $ and 
$\sum_{j=1}^{k}\beta_{j}^2 = \left|P_{S_1^\bot}\left(\overrightarrow{v^N}/|\overrightarrow{v^N}| \right)\right|^2$. 
First, let us assume that $\sum_{j=1}^{k}\beta_{j}^2\neq 0$ and let $(\alpha_1^\ast,\ldots,\alpha_{n-1}^\ast) $ be a solution of the problem above.
Using well-known necessary optimality conditions for constrained minimization, it follows that there exists a (Lagrange) multiplier $\lambda\in R$ satisfying the KKT condition
\[
 \nabla p(\alpha_1^\ast,\ldots,\alpha_{k}^\ast) +\lambda \nabla q (\alpha_1^\ast,\ldots,\alpha_{k}^\ast) = 0,\quad
 \sum_{j=1}^{k}(\alpha_j^\ast)^2= 1.
 \]
 That is
 \[
 \beta_{j}+2\lambda \alpha_j^\ast= 0\quad j=1,\ldots,k,\quad
 \sum_{j=1}^{k}(\alpha_j^\ast)^2=1.
\]
In view of the assumptions and the above conditions it follows that $\lambda\neq 0$ and it holds
\[
\sum_{j=1}^{k}\frac{\beta_{i}^2}{4\lambda^2}= \sum_{j=1}^{k}(\alpha_j^\ast)^2=1
\]
Thus, $\lambda^2 = \sum_{j=1}^{k} \beta_{j}^2/4$, hence it follows
$\lambda=\sqrt{\sum_{j=1}^{k} \beta_{j}^2}/2$ or $\lambda=-\sqrt{\sum_{j=1}^{k} \beta_{j}^2}/2$. 
 Since the problem has at least one solution, and it holds 
 $ \alpha_j^\ast = -\beta_{j}/2\lambda$ for $j=1,\ldots,k$, it follows from the definition of the objective function in Problem \eqref{eq:REFORM} that the minimum is achieved when $\lambda <0$. Therefore, it follows that
\[ x^\ast=\sum_{j=1}^{k}\alpha^\ast_j w_j = 
 \sum_{j=1}^{k}\frac{ \beta_{j}}{ \sqrt{\sum_{j=1}^{k} \beta_{j}^2}}w_j
 =\frac{ P_{S_1^\bot}\left(\overrightarrow{v^N}/|\overrightarrow{v^N}|\right)}{ \left|P_{S_1^\bot}\left(\overrightarrow{v^N}/|\overrightarrow{v^N}|\right)\right|}
 \]
 {is a solution of Problem \eqref{eq:EXIST_B1}.}
Note that given an orthonormal base of $<b>^\bot$, calculating the vector $x^\ast$ above is trivial. 
 To end the argument, note that $\sum_{j=1}^{k}\beta_{j}^2 =0$ implies that 
 $$
 \left< \frac{\overrightarrow{v^N}}{|\overrightarrow{v^N}|},\frac{\overrightarrow{v^N}}{|\overrightarrow{v^N}|}-\frac{\overrightarrow{v^j}}{|\overrightarrow{v^j}|}\right> =0 \label{eq:exist.a} \quad j=1,\ldots,N-1,
 $$
From the above relation, it follows that $d(\overrightarrow{v^N}/|\overrightarrow{v^N},\overrightarrow{v^j}/|\overrightarrow{v^j})=d(\overrightarrow{v^N}/|\overrightarrow{v^N},\overrightarrow{v^N}/|\overrightarrow{v^N})=0$ for $j=1,\ldots,N-1$, 
{which,  in view of \eqref{eq:int.1}, implies that $\overrightarrow{v^j}/|\overrightarrow{v^j}|=\overrightarrow{v^N}/|\overrightarrow{v^N}|$ for $j=1,\ldots,N-1$, and contradicts our initial assumption.}
    \end{proof}

    Proof of Proposition \ref{prop:P2.related.refor}.
    \begin{proof}
 The set of solutions of the linear  system in \eqref{prob2.refor} is $S=\{x_0\}+S_0$
where $S_0=\left\{\hat x\in \R^n\, |\, V\hat x=0\right\}\neq\{0\} $.
Thus, in order to solve problem \eqref{prob2.refor} we need to find a vector $x =x_0+\hat x$ with $\hat x\in S_0$ and
such that $\norm{x}=1$.
Note that, in  this case, it holds
\[
1= \norm{x_0+\hat x}\geq \min_{x\in S_0}\norm{x_0+ x}=\norm{x_0-P_{S_0}(x_0)}
 \]
 where $P_{S_0}(x_0) =  \left(I_d-V^t\left(VV^t\right)^{-1}V\right)x_0 = x_0-V^t\left(VV^t\right)^{-1}w$. 
 Thus, it follows that the problem \eqref{prob2.refor}  has a solution only if
  $\norm{x_0-P_{S_0}(x_0)}=\left\|V^t\left(VV^t\right)^{-1}w
\right\|\leq 1$.
Next, we will prove that this condition is also sufficient. 
Note that, if it  holds,  we can find a solution to the problem \eqref{prob2.refor} as follows:   $ {x^\ast = x_0-P_{S_0}(x_0)+t^\ast x_1}\in S$, where $0\neq x_1\in S_0$ and
$t^\ast $ satisfies $\norm{x_0 - P_{S_0}(x_0)+t^\ast 
 x_1}^2=1$.
Simple calculations show that
\begin{align*}
  p(t)&=\norm{x_0 - P_{S_0}(x_0)+tx_1}^2=
  t^2\norm{x_1}^2-t2\left<x_0-P_{S_0}(x_0),x_1\right> +\norm{x_0-P_{S_0}(x_0)}^2\\
   &=  t^2\norm{x_1}^2+\norm{x_0-P_{S_0}(x_0)}^2
   =t^2\norm{x_1}^2+\left\|V^t\left(VV^t\right)^{-1}w
\right\|^2,
 \end{align*}
Hence, the  equation $p(t)=1$ has the trivial solution 
 $ t^\ast = \sqrt{\left(1-\left\|V^t\left(VV^t\right)^{-1}w
\right\|^2\right)/\norm{x_1}^2}$, and the claim follows.
As a side note, notice that taking $x_1= P_{S_0}(x_0)$, when $P_{S_0}(x_0)\neq 0$, turns  the expression above for  the solution  to problem \eqref{prob2.refor} into
\[
x^\ast=x_0-\left(1-\sqrt{\frac{\te 1-\left\|V^t\left(VV^t\right)^{-1}w
\right\|^2}{\te \norm{P_{S_0}}^2}}\right)P_{S_0}(x_0)
\]
\end{proof}

\section{Results for the KNN classifier}

\begin{table*}[htbp]
  \caption{Results of KNN for probing tasks using different embedding methods.}
  \label{tab:Content_Results_KNN}
  \vskip 0.15in
  \tiny  
  \centering
  \begin{sc}
    \begin{tabular}{c|rr|rr|rr|rr|rr|r|rr}
\toprule
     Embedding&  \multicolumn{2}{c|}{fasttext}  &  \multicolumn{2}{c|}{glove} &  \multicolumn{2}{c|}{word2vec} &   \multicolumn{2}{c|}{BERT} &  \multicolumn{2}{c|}{RoBERTa} &    USE & \multicolumn{2}{c}{total} \\ \hline
Benchmark&       BOV &       OSE &    BOV &    OSE &      BOV &      OSE &    BOV &    OSE &      BOV &      OSE &  &      BOV &      OSE   \\ \hline \hline

Length & 34.98  & 39.35  & 34.59  & 28.05  & 33.99  & 33.73  & 38.45  & 42.10  & {\bf{43.37}}  & 34.61  & 43.04  &3 &2 \\
WordContent & 30.33  & 44.14  & 22.96  & {\bf{50.31}}  & 18.95  & 48.50 & 16.50 & 18.62  & 4.92  & 6.59 & 44.41  &0 &5 \\
Depth & 25.05  & 26.47  & 23.58  & 23.73  & 23.99  & 24.10  & {\bf{27.40}}  & 27.15  & 22.21  & 22.52  & 23.09   &1 &4 \\
TopConstituents & 36.74 & 50.63 & 32.00 & 52.08 & 35.39 & 39.21 & 48.28  & {\bf{59.65}} & 24.56  & 27.68 & 49.09 &0 &5 \\
BigramShift & 49.18  & 49.70  & 49.83 & 49.83 & 48.96 & 48.58 & 71.86  & {\bf{73.52}}  & 50.27  & 50.79  & 52.68 &1 &3 \\
Tense & 77.62  & 78.43  & 75.76  & 76.47  & 74.11  & 75.75  & 83.49  & {\bf{83.54}}  & 57.44  & 58.05  & 69.90   &0 &5 \\
SubjNumber & 70.89  & {\bf{74.67}}  & 71.28  & 71.14 & 70.99 & 74.19 & 69.52 & 70.42 & 60.39  & 60.40  & 64.10   &1 &4 \\
ObjNumber & {\bf{67.11}}  & 66.77  & 66.47  & 62.73 & 63.81 & 64.24 & 65.35  & 64.66  & 58.27  & 57.17  & 59.14  &4 &1 \\
OddManOut & 51.29  & 51.51  & 50.97  & 47.01 & 49.78 & 49.41 & 56.47  & {\bf{56.86}}  & 49.85  & 49.78  & 50.16  &3 &2 \\
CoordinationInversion &50.37 & 52.67 & 51.18 & 51.76 & 51.56 & 51.82 & 55.95 &{\bf{58.65}} &51.95 &53.37 &52.49  &0 &5 \\ \hline
Total &1   &9      &4  &5     &3    &7   &2   &8  &3  &7  &  &13 & 36  \\ 

\bottomrule
    \end{tabular}
  \end{sc}
  \vskip -0.1in
\end{table*}

\begin{table*}[htbp]
   \caption{Results of KNN for sentence classification and natural language inference tasks. }
   \label{tab:Semantic_Results_KNN}
   \vskip 0.15in
 \tiny
 \centering
   \begin{sc}
    \begin{tabular}{c|rr|rr|rr|rr|rr|r|rr}
\toprule
     Embedding&  \multicolumn{2}{c|}{fasttext}  &  \multicolumn{2}{c|}{glove} &  \multicolumn{2}{c|}{word2vec} &   \multicolumn{2}{c|}{BERT} &  \multicolumn{2}{c|}{RoBERTa} &    USE & \multicolumn{2}{c}{total} \\ \hline
Benchmark&       BOV &       OSE &    BOV &    OSE &      BOV &      OSE &    BOV &    OSE  &      BOV &      OSE &  &      BOV &      OSE     \\ \hline \hline
MR & 69.38  & 71.29  & 68.93  & 69.48  & 68.05  & 68.72  & {\bf{73.78}}  & 71.40  & 53.24  & 52.32  & 71.21 &2 & 3\\
CR & 74.36  & 74.46  & 74.17  & 71.42  & 73.54  & 72.98  & {\bf{78.62}}  & 76.37  & 60.03  & 62.54  & 76.61 &3 & 2 \\
SUBJ & 89.48  & 90.74  & 89.16  & 86.97  & 88.12  & 89.23  & {\bf{93.20}}  & 92.66  & 64.68  & 59.66  & 91.38 &3 &2 \\
MPQA & 85.55  & 85.32  & 84.13  & 84.63  & 85.24  & 85.40  & 75.24  & 75.08  & 68.63  & 67.72  & {\bf{86.43}} &3 &2 \\
TREC & 75.60  & {\bf{86.20}}  & 74.40  & 76.40  & 76.80  & 84.20  & 83.80  & 86.00  & 56.60  & 56.20  & 85.20 &1 &4 \\
SST2 & 71.33  & 71.22  & 70.57  & 71.55  & 69.63  & 70.62  & {\bf{76.39}}  & 73.92  & 52.72  & 55.08  & 73.20 &2 &3 \\
SST5 & 31.22  & 35.07  & 33.71  & 31.81  & 33.85  & 32.58  & 35.07  & 35.38  & 23.26  & 22.94  & {\bf{36.02}} &3 &2 \\
MRPC & 69.97  & 68.58  & 69.80  & 66.78  & 64.00  & 62.72  & {\bf{70.90}}  & 67.48  & 66.14  & 67.01  & 58.61 &4 &1 \\
SICKEntailment & 76.09 & 77.80 & 66.35 & 72.30 & 76.74 & 77.59 & 66.41 & 72.09 & 50.80 & 57.46 & {\bf{78.49}} &0 &5 \\ \hline
Total & 3  & 6     & 4 & 5    & 3 & 6   & 6   & 3  & 5  & 4 &     & 21  & 24 \\ 
\bottomrule
    \end{tabular}
   \end{sc}
   \vskip -0.1in
\end{table*}

\end{document}